\documentclass[10pt,twocolumn,letterpaper]{article}

\usepackage[pagenumbers]{cvpr} 

\usepackage{graphicx}
\usepackage{amsmath}
\usepackage{amssymb}
\usepackage{booktabs}

\usepackage[pagebackref,breaklinks,colorlinks]{hyperref}

\usepackage[capitalize]{cleveref}
\crefname{section}{Sec.}{Secs.}
\Crefname{section}{Section}{Sections}
\Crefname{table}{Table}{Tables}
\crefname{table}{Tab.}{Tabs.}

\usepackage{wrapfig}
\usepackage{multirow, makecell}
\usepackage{url}            
\usepackage{booktabs}       
\usepackage{amsfonts}       
\usepackage{nicefrac}       
\usepackage{microtype}      
\usepackage{xcolor}         
\usepackage{url}         

\usepackage{amsmath,amsfonts,bm}

\def\eqref#1{equation~\ref{#1}}

\def\1{\bm{1}}

\def\rva{{\mathbf{a}}}

\def\rvx{{\mathbf{x}}}
\def\rvy{{\mathbf{y}}}

\DeclareMathAlphabet{\mathsfit}{\encodingdefault}{\sfdefault}{m}{sl}
\SetMathAlphabet{\mathsfit}{bold}{\encodingdefault}{\sfdefault}{bx}{n}

\newcommand{\pdata}{p_{\rm{data}}}

\usepackage[sort, numbers]{natbib}

\def\cvprPaperID{43} 
\def\confName{CVPR}
\def\confYear{2023}

\begin{document}

\title{Visual Chain-of-Thought Diffusion Models}

\author{William Harvey\\
University of British Columbia\\
{\tt\small wsgh@cs.ubc.ca}
\and
Frank Wood\\
University of British Columbia\\
{\tt\small fwood@cs.ubc.ca}
}
\maketitle

\begin{abstract}
   Recent progress with conditional image diffusion models has been stunning, and this holds true whether we are speaking about models conditioned on a text description, a scene layout, or a sketch. Unconditional image diffusion models are also improving but lag behind, as do diffusion models which are conditioned on lower-dimensional features like class labels. We propose to close the gap between conditional and unconditional models using a two-stage sampling procedure. In the first stage we sample an embedding describing the semantic content of the image. In the second stage we sample the image conditioned on this embedding and then discard the embedding. Doing so lets us leverage the power of conditional diffusion models on the unconditional generation task, which we show improves FID by $25 - 50\%$ compared to standard unconditional generation.
\end{abstract}

\section{Introduction}
\label{sec:intro}
\begin{figure}
    \centering
    \includegraphics[width=0.49\columnwidth]{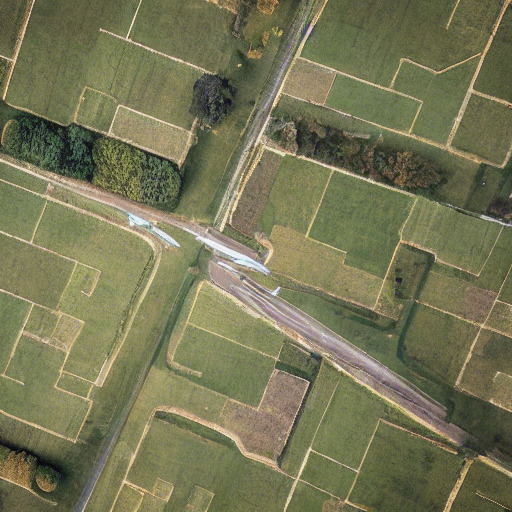}
    \includegraphics[width=0.49\columnwidth]{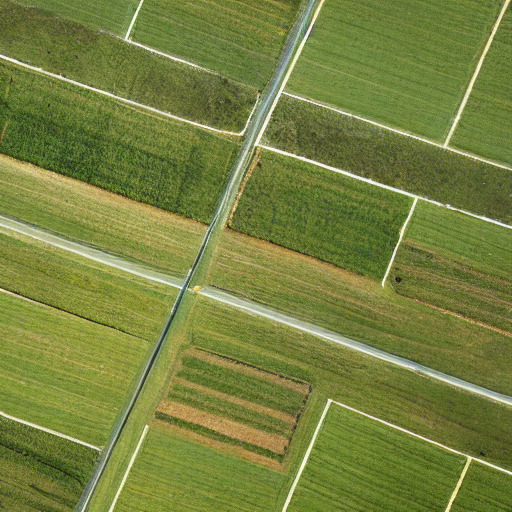}
    \caption{\textbf{Left:} Output from Stable Diffusion~\citep{rombach2022high} prompted to produce ``aerial photography''. \textbf{Right:} Using a more detailed prompt\protect\footnotemark with the same random seed removes the ``smudged'' road artifact that appears on the left. VCDM builds on this observation.}
    \label{fig:stable-diffusion-example}
\end{figure}

Recent text-to-image diffusion generative models (DGMs) have exhibited stunning sample quality~\citep{saharia2022photorealistic} to the point that they are now being used to create art~\citep{oppenlaender2022creativity}. 
Further work has explored conditioning on scene layouts~\citep{zhang2023adding}, segmentation masks~\citep{zhang2023adding,hu2022self}, or the appearance of a particular object \citep{ma2023unified}. We broadly lump these methods together as ``conditional'' DGMs to contrast them with ``unconditional'' image DGMs which sample an image without dependence on text or any other information.
Relative to unconditional DGMs, conditional DGMs typically produce more realistic samples~\citep{ho2022classifier,bao2022conditional,hu2022self} and work better with few sampling steps~\citep{meng2022distillation}. Furthermore our results suggest that sample realism grows with ``how much'' information the DGM is conditioned on: as hinted at in \cref{fig:stable-diffusion-example} an image is likely to be more realistic if conditioned on being ``an aerial photograph of a road between green fields'' than if it is if simply conditioned on being ``an aerial photograph.''

This gap in performance is problematic. In the spirit of this workshop, imagine you have been tasked with sampling a dataset of synthetic aerial photos which will be used to train a computer vision system. A researcher doing so would currently have to either (a) make up a scene description before generating each dataset image, and ensure these cover the entirety of the desired distribution, or (b) accept the inferior image quality gleaned by conditioning just on each image being ``an aerial photograph''. \footnotetext{We used the prompt ``Aerial photography of a patchwork of small green fields separated by brown dirt tracks between them. A large tarmac road passes through the scene from left to right.''}

To close this gap, we take inspiration from ``chain of thought'' reasoning~\citep{wei2022chain} in large language models (LLMs). Consider using an LLM to answer a puzzle: \textit{Roger has 5 tennis balls. He buys 2 more cans of tennis balls. Each can has 3 tennis balls. How many tennis balls does he have now?}
If the LLM is prompted to directly state the answer, it must perform all reasoning and computation in a single step. If instead it is prompted to explain its reasoning as it computes an answer, it can first conclude that the answer is given by the expression $5+2\times3$, and then output an answer \textit{conditioned} on it arising from such an expression. Printing this expression in an intermediate step dramatically improves accuracy~\citep{wei2022chain}.

Let us imagine an image generative model along these lines. When prompted to sample ``an aerial photograph'', it may start by sampling a more detailed description: ``an aerial photograph of a patchwork of small green fields [...]''. 
Given this detailed description, it can leverage the full power of a conditional DGM to generate a high-quality image. Our approach follows these lines but, instead of operating on language, our intermediate space consists of a semantically-meaningful embedding from a pretrained CLIP embedder~\citep{radford2021learning}.

Specifically we train a DGM to model the distribution of CLIP embeddings of images in our dataset. From this we achieve improved unconditional image generation by first sampling a CLIP embedding  and then feeding this CLIP embedding into a conditional image DGM.
Note that, while this technique is related to text-conditional image generation, we are instead applying it to improved {\em unconditional} image generation. We call the resulting model a Visual Chain-of-Thought Diffusion Model (VCDM) and release code at \url{https://github.com/plai-group/vcdm}.

\section{Background}
\paragraph{Conditional DGMs}
We provide a high-level overview of conditional DGMs that is sufficient to understand our contributions, referring to \citet{karras2022elucidating} for a more complete description and derivation. A conditional image DGM~\citep{tashiro2021csdi} samples an image $\rvx$ given a conditioning input $\rvy$, where $\rvy$ can be, for example, a class label, a text description, or both of these in a tuple. We can recover an unconditional DGM by setting $\rvy$ to a null variable in the below. Given a dataset of $(\rvx,\rvy)$ pairs sampled from $\pdata(\cdot,\cdot)$, a conditional DGM $p_\theta(\rvx|\rvy)$ is fit to approximate $\pdata(\rvx|\rvy)$. It is parameterized by a neural network $\hat{\rvx}_\theta(\cdot)$ trained to optimize
\begin{align}\label{eq:diffusion-loss}
    \mathbb{E}_{u(\sigma)p_\sigma(\rvx_\sigma|\rvx,\sigma)\pdata(\rvx,\rvy)} \left[ \lambda(\sigma) \lvert\lvert \rvx - \hat{\rvx}_\theta(\rvx_\sigma, \rvy, \sigma) \rvert\rvert^2 \right]
\end{align}
where $\rvx_\sigma \sim p_\sigma(\cdot|\rvx,\sigma)$ is a copy of $\rvx$ corrupted by Gaussian noise with standard deviation $\sigma$; $u(\sigma)$ is a broad distribution over noise standard deviations; and $\lambda(\sigma)$ is a weighting function. During inference, samples from $p_\theta(\rvx|\rvy)$ are drawn via a stochastic differential equation with dynamics dependent on $\hat{\rvx}_\theta(\cdot)$.

\begin{figure}[t]
    \centering
    \includegraphics[width=0.49\columnwidth]{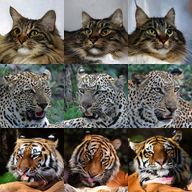}
    \includegraphics[width=0.49\columnwidth]{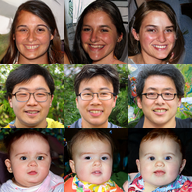}
    \caption{CLIP-conditional samples on AFHQ and FFHQ. Each row shows three samples conditioned on the same CLIP embedding.}
    \label{fig:samples}
\end{figure}

\begin{figure}[t]
    \centering
    \includegraphics[width=\columnwidth]{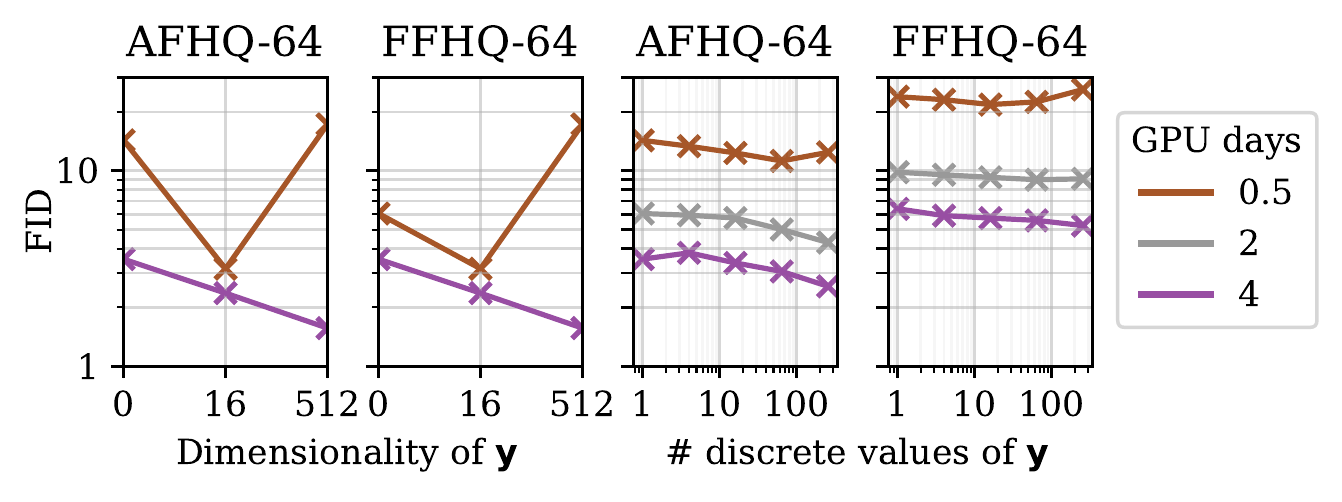}
    \vspace{-4mm}
    \caption{FID versus dimensionality of $\rvy$ on AFHQ~\citep{choi2020stargan} and FFHQ~\citep{karras2018style}. With small training budgets (brown line), it is harmful when $\rvy$ is too informative. With larger training budgets (purple line), it is helpful to make $\rvy$ much more high dimensional.}
    \label{fig:fid-vs-ncomp}
\end{figure}

\paragraph{CLIP embeddings}
CLIP (contrastive language-image pre-training)~\cite{radford2021learning} consists of two neural networks, an image embedder $e_i(\cdot)$ and a text embedder $e_t(\cdot)$, trained on a large captioned-image dataset. Given an image $\rvx$ and a caption $\rvy$, the training objective encourages the cosine similarity between $e_i(\rvx)$ and $e_t(\rvy)$ to be large if $\rvx$ and $\rvy$ are a matching image-caption pair and small if not.
The image embedder therefore learns to map from an image to a semantically-meaningful embedding capturing any features that may be included in a caption. We use a CLIP image embedder with the ViT-B/32 architecture and weights released by \citet{radford2021learning}. We can visualize the information captured by the CLIP embedding by showing the distribution of images produced by our conditional DGM given a single CLIP embedding; see \cref{fig:samples}.

\section{Conditional vs. unconditional DGMs}
\paragraph{What does it mean to say that conditional DGMs beat unconditional DGMs?} A standard procedure to evaluate unconditional DGMs is to start by sampling a set of $N$ images independently from the model: ${\rvx^{(1)},\ldots,\rvx^{(N)} \sim p_\theta(\cdot)}$. We can then compute the Fr\'echet Inception distance (FID)~\citep{heusel2017gans} between this set and the dataset. If the generative model matches the data distribution well, the FID will be low.
For conditional DGMs the standard procedure has one extra step: we first independently sample ${\rvy^{(1)},\ldots,\rvy^{(N)} \sim \pdata(\cdot)}$. We then sample each image given the corresponding $\rvy^{(i)}$ as ${\rvx^{(i)} \sim p_\theta(\cdot|\rvy^{(i)})}$. 
Then, as in the unconditional case, we compute the FID between the set of images $\rvx_1,\ldots,\rvx_N$ and the dataset, without reference to $\rvy_1,\ldots,\rvy_N$. Even though it does not measure alignment between $\rvx, \rvy$ pairs, conditional DGMs beat comparable unconditional DGMs on this metric in many settings: class-conditional CIFAR-10 generation~\citep{karras2022elucidating}, segmentation-conditional generation~\cite{hu2022self}, or bounding box-conditional generation~\citep{hu2022self}.


\begin{figure*}[ht]
    \centering
    \includegraphics[width=\textwidth]{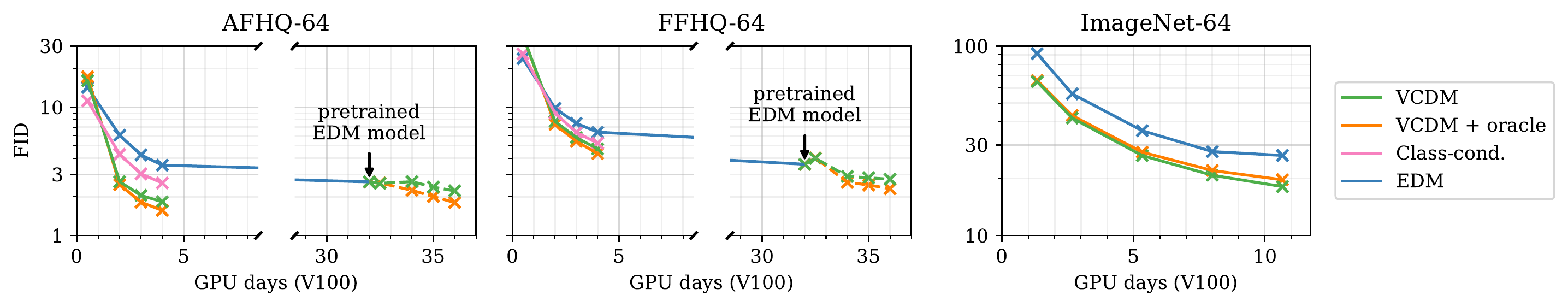}
    \vspace{-5mm}
    \caption{FID throughout training. We show results for each method trained from scratch and, on AFHQ and FFHQ, for finetuning a pretrained EDM model (which was trained for the equivalent of 32 GPU days). VCDM quickly outperforms EDM when trained from scratch and quickly improves on the pretrained model when used for finetuning.}
    \vspace{-2mm}
    \label{fig:fid_vs_training}
\end{figure*}

\paragraph{Why do conditional DGMs beat unconditional DGMs?}

Conditional DGMS ``see'' more data during training than their unconditional counterparts because updates involves $\rvy$ as well as $\rvx$. \citet{bao2022conditional,hu2022self} prove that this is not the sole reason for their successes because the effect holds up even when $\rvy$ is derived from an unconditional dataset through self-supervised learning.
To our knowledge, the best explanation for their success is, as stated by \citet{bao2022conditional}, that conditional distributions typically have ``fewer modes and [are] easier to fit than the original data distribution.''

\paragraph{When do conditional DGMs beat unconditional DGMS?}
%
We present results in \cref{fig:fid-vs-ncomp} to answer this question. We show FID scores for conditional DGMs trained to condition on embeddings of varying information content. 
We produce $\rvy$ by starting from the CLIP embedding of each image in our dataset and using either principal component analysis to reduce their dimensionality (left two panels) or K-means clustering to discretize them (right two panels)~\citep{hu2022self}.
%
We see that, given a small training budget, it is best to condition on little information. With a larger training budget, performance appears to improve steadily as the dimensionality of $\mathbf{y}$ is expanded. We hypothesize that \textbf{(1)} conditioning on higher-dimensional $\mathbf{y}$ slows down training because it means that points close to any given value of $\mathbf{y}$ will be seen less frequently and $\textbf{(2)}$ with a large enough compute budget, any $\mathbf{y}$ correlated with $\mathbf{x}$ will be useful to condition on. This suggests that, as compute budgets grow, making unconditional DGM performance match conditional DGM performance will be increasingly useful.
%


\section{Method} \label{sec:method}
We have established that conditioning on CLIP embeddings improves DGMs. We now introduce VCDM which leverages this phenomenon to benefit the unconditional setting (in which the user does not wish to specify any input to condition on) and the ``lightly-conditional'' setting in which the input is low-dimensional, e.g. a class-label. We will denote any such additional input $\rva$ (letting $\rva$ be a null variable in the unconditional setting) and from now on always use $\rvy := e_i(\rvx)$ to refer to a CLIP embedding. VCDM approximates the target distribution $\pdata(\rvx|\rva)$ as
%
%
\begin{align} \label{eq:no-a}
    \pdata(\rvx|\rva) &= \mathbb{E}_{\pdata(\rvy|\rva)} \left[ \pdata(\rvx|\rvy,\rva) \right] \\
    &\approx \mathbb{E}_{p_\phi(\rvy|\rva)} \left[ p_\theta(\rvx|\rvy,\rva) \right] 
\end{align}
where $p_\phi(\rvy|\rva)$ is a second DGM modeling the CLIP embeddings. We can sample from this distribution by sampling $\rvy\sim p_\phi(\cdot|\rva)$ and then leveraging the conditional image DGM to sample $\rvx \sim p_\theta(\cdot|\rvy,\rva)$ before discarding $\rvy$.
%
From now on we will call $p_\theta(\rvx|\rvy,\rva)$ the \textit{conditional image model} and $p_\phi(\rvy|\rva)$ the \textit{auxiliary model}. In our experiments the auxiliary model uses a small architecture relative to the conditional image model and so adds little extra cost.\footnote{For our ImageNet experiments, sampling from our auxiliary model takes 35ms per batch item. Sampling from our image model takes 862ms and so VCDM has inference time only $4\%$ greater than our baselines.}

\paragraph{Auxiliary model}
Our auxiliary model is a conditional DGM targeting $\pdata(\rvy|\rva)$, where $\rvy$ is a 512-dimensional CLIP embedding. We follow the architectural choice of \citet{ramesh2022hierarchical} and use a DGM with a transformer architecture. It takes as input a series of 512-dimensional input tokens: an embedding of $\sigma$; an embedding of $\rva$ if this is not null; an embedding of $\rva_\sigma$; and a learned query. These are passed through six transformer layers and then the output corresponding to the learned query token is used as the output. Like \citet{ramesh2022hierarchical}, we parameterize the DGM to output an estimate of the denoised $\rva$ instead of estimating the added noise as is more common in the diffusion literature.
On AFHQ and FFHQ we find that data augmentation is helpful to prevent the auxiliary model overfitting. We perform augmentations (including rotation, flipping and color jitter) in image space and feed the augmented image through $e_i(\cdot)$ to obtain an augmented CLIP embedding. Following \citet{karras2022elucidating}, we pass a label describing the augmentation into the transformer as an additional input token so that we can condition on there being no augmentation at test-time.

\paragraph{Conditional image model}
Our diffusion process hyperparameters and samplers build on those of \citet{karras2022elucidating}.  For AFHQ and FFHQ, we use the U-Net architecture originally proposed by \citet{song2020score}. For ImageNet, we use the slightly larger U-Net architecture originally proposed by \citet{dhariwal2021diffusion}. We match the data augmentation scheme to be the same as that of \citet{karras2022elucidating} on each dataset. There are established conditional variants of both architectures we use~\citep{dhariwal2021diffusion,karras2022elucidating}, both of which incorporate $\rvy$ via a learned linear projection that is added to the embedding of the noise standard deviation $\sigma$. Our conditional image model needs to additionally incorporate $\rva$; we can do so by simply concatenating it to $\rvy$ and learning a projection for the resulting vector.



\section{Experiments}
We experiment on three datasets: AFHQ~\citep{choi2020stargan}, FFHQ~\citep{karras2018style} and ImageNet~\citep{deng2009imagenet}, all at $64\times64$ resolution. We target unconditional generation for AFHQ and FFHQ, and class-conditional generation for ImageNet. As well as training networks from scratch on each dataset, we report results with the model checkpoints released by \citet{karras2022elucidating} on AFHQ and FFHQ, which we finetune to be conditional on the CLIP embedding. To do so, we simply add a learnable linear projection of the CLIP embedding and initialize its weights to zero.
\Cref{fig:fid_vs_training} reports the FID on each setting and dataset throughout the training of the conditional image model.\footnote{Each FID is estimated using $20\,000$ images, each sampled with the SDE solver proposed by \citet{karras2022elucidating} using 40 steps, $S_\text{churn}=50$, $S_\text{noise}=1.007$, and other parameters set to their default values.} 
In each case, the auxiliary model is trained for one day on one V100 GPU. We compare VCDM to three other approaches: \textbf{EDM}~\citep{karras2018style} is a standard DGM directly modeling $\pdata(\rvx|\rva)$. \textbf{VCDM with oracle} uses our conditional image model but uses the ground-truth $\rvy$ for each test $\rva$ instead of sampling from the learned auxiliary model, i.e. it is the performance that VCDM would achieve with a perfect auxiliary model. \textbf{Class-cond} is an ablation of VCDM that applies to unconditional tasks where $\rva$ is null. It uses discrete $\rvy$ (as on the right of \cref{fig:fid-vs-ncomp}) so that $\pdata(\rvy|\rva)=\pdata(\rvy)$ is a simple categorical distribution which we can sample from exactly, but we see that it is outperformed by VCDM.

VCDM consistently outperforms unconditional generation after 1-2 GPU-days and this performance gap continues for as long as we train the networks. Comparing VCDM's performance with and without the oracle we see that they are close. For networks trained from scratch we show in \cref{tab:results} that VCDM always has an improvement over EDM at least $80\%$ as large as that of VCDM with an oracle, indicating that $p_\phi(\rvy|\rva)$ is a good approximation of $\pdata(\rvy|\rva)$. We can therefore leverage almost the full power of conditional DGMs for unconditional sampling. 

Sampling each CLIP embedding from our auxiliary model takes only 35ms. Sampling from our image model takes 862ms and so there is only a $4\%$ increase in inference time when using VCDM vs EDM.

\begin{table} \label{tab:results}
\centering
\caption{Final FID score for the models we train from scratch and a comparison of their improvements over EDM.}
\begin{tabular}{lccc}
\midrule
Dataset & AFHQ & FFHQ & ImageNet \\
\midrule
$\mathbf{y}$ & null & null & class label \\
\midrule
VCDM & $1.83$ & $4.73$ & $18.1$ \\
VCDM + oracle & $1.57$ & $4.35$ & $19.7$ \\
Class-cond. & $2.56$ & $5.24$ & - \\
EDM & $3.53$ & $6.39$ & $26.5$ \\
\midrule
\begin{tabular}{@{}l@{}}Improv. w/ VCDM\end{tabular} & $48.2\%$ & $26.0\%$ & $31.5\%$ \\
\begin{tabular}{@{}l@{}}Improv. w/ oracle\end{tabular} & $55.6\%$ & $31.9\%$ & $25.6\%$ \\
\midrule
$\frac{\text{Improv. w/ VCDM}}{\text{Improv. w/ oracle}}$ & $86.6\%$ & $81.3\%$ & $123\%$ \\
\end{tabular}
\end{table}



\section{Related work}
Several existing image generative models leverage CLIP embeddings for better text-conditional generation~\citep{nichol2021glide,ramesh2022hierarchical}. We differ by suggesting that CLIP embeddings are not only useful for text-conditioning, but also as a general tool to improve the realism of generated images. We demonstrate this for unconditional and class-conditional generation. Our work takes inspiration from \citet{weilbach2022graphically}, 
who show improved performance in various approximate inference settings by modeling problem-specific auxiliary variables (like $\rvy$) in addition to the variables of interest ($\rvx$) and observed variables ($\rva$). We apply these techniques to the image domain and incorporate pretrained CLIP embedders to obtain auxiliary variables. 
VCDM also relates to methods which perform diffusion in a learned latent space~\citep{rombach2022high}: our auxiliary model $p_\phi(\rvy|\rva)$ is analogous to a ``prior'' in a latent space and our conditional image model $p_\theta(\rvx|\rva,\rvy)$ to a ``decoder'' Such methods typically use a near-deterministic decoder and so their latent variables must summarize all information about the image. Our conditional DGM decoder on the other hand will function reasonably however little information is stored in $\rvy$ and so VCDM provides an additional degree of freedom in terms of what to store. This is an interesting design space for future exploration. Classifier~\citep{song2020score} and classifier-free guidance~\citep{ho2022classifier} are two alternative methods for conditional sampling from DGMs. Both have a ``guidance strength'' hyperparameter to trade fidelity to $\pdata(\rvx|\rvy)$ against measures of alignment between $\rvx$ and $\rvy$. A possible extension to VCDM could parameterize $p_\theta(\rvx|\rvy,\rva)$ with either of them.

\section{Discussion and conclusion}
We have presented VCDM, a method for unconditional or lightly-conditional image generation which harnesses the impressive performance of conditional DGMs. A massive unexplored design space remains: there are almost certainly more useful quantities that we could condition on than CLIP embeddings. It may also help to condition on multiple quantities, or ``chain'' a series of conditional DGMs together. An alternative direction is to simplify VCDM's architecture by, for example, learning a single diffusion model over the joint space of $\rvx$ and $\rvy$ instead of generating them sequentially. A drawback of VCDM is that it relies on the availability of a pretrained CLIP embedder. While this is freely available for natural images, it could be a barrier to other applications; an alternative would be to explore the self-supervised representations used by \citet{bao2022conditional,hu2022self}.

\section*{Acknowledgments}
We acknowledge the support of the Natural Sciences and Engineering Research Council of Canada (NSERC), the Canada CIFAR AI Chairs Program, and the Intel Parallel Computing Centers program. This material is based upon work supported by the United States Air Force Research Laboratory (AFRL) under the Defense Advanced Research Projects Agency (DARPA) Data Driven Discovery Models (D3M) program (Contract No. FA8750-19-2-0222) and Learning with Less Labels (LwLL) program (Contract No.FA8750-19-C-0515). Additional support was provided by UBC's Composites Research Network (CRN), Data Science Institute (DSI) and Support for Teams to Advance Interdisciplinary Research (STAIR) Grants. This research was enabled in part by technical support and computational resources provided by WestGrid (https://www.westgrid.ca/) and Compute Canada (www.computecanada.ca).

{\small
\bibliographystyle{ieee_fullname_natbib}
\bibliography{references}
}

\end{document}